\documentclass[sigconf]{acmart}

\usepackage{multirow}
\usepackage{subfigure}
\usepackage{comment}
\usepackage{amsmath}
\let\vec\mathbf
\usepackage{enumitem}% http://ctan.org/pkg/enumitem
\usepackage{color}
\usepackage{bm}

\usepackage{etoolbox}
\makeatletter
\patchcmd{\maketitle}{\@copyrightspace}{}{}{}
\makeatother

\newcommand{\bhline}[1]{\noalign{\hrule height #1}}
\AtBeginDocument{%
  \providecommand\BibTeX{{%
    \normalfont B\kern-0.5em{\scshape i\kern-0.25em b}\kern-0.8em\TeX}}}

\begin{document}
\pagestyle{plain}
\settopmatter{printacmref=false} % Removes citation information below abstract
\renewcommand\footnotetextcopyrightpermission[1]{} % removes footnote with conference information in first column
\pagestyle{plain} % removes running headers

\title{A Picture May Be Worth a Hundred Words \\for Visual Question Answering}


\author{Yusuke Hirota}
\affiliation{%
  \institution{Osaka University}
  \city{Osaka}
    \country{Japan}
}
\email{y-hirota@ist.osaka-u.ac.jp}

\author{Noa Garcia}
\affiliation{%
  \institution{Osaka University}
  \city{Osaka}
  \country{Japan}
  }
\email{noagarcia@ids.osaka-u.ac.jp}

\author{Mayu Otani}
\affiliation{%
  \institution{CyberAgent, Inc.}
  \city{Tokyo}
  \country{Japan}
}
\email{otani\_mayu@cyberagent.co.jp}

\author{Chenhui Chu}
\affiliation{%
 \institution{Kyoto University}
 \city{Kyoto}
 \country{Japan}
}
 \email{chu@i.kyoto-u.ac.jp}

\author{Yuta Nakashima}
\affiliation{%
  \institution{Osaka University}
  \city{Osaka}
  \country{Japan}
  }
 \email{n-yuta@ids.osaka-u.ac.jp}

\author{Ittetsu Taniguchi}
\affiliation{%
  \institution{Osaka University}
  \city{Osaka}
  \country{Japan}
}
\email{i-tanigu@ist.osaka-u.ac.jp}

\author{Takao Onoye}
\affiliation{%
  \institution{Osaka University}
  \city{Osaka}
  \country{Japan}
}
\email{onoye@ist.osaka-u.ac.jp}

\renewcommand{\shortauthors}{Hirota, et al.}

\begin{abstract}
How far can we go with textual representations for understanding pictures? In image understanding, it is essential to use concise but detailed image representations. Deep visual features extracted by vision models, such as Faster R-CNN, are prevailing used in multiple tasks, and especially in visual question answering (VQA). However, conventional deep visual features may struggle to convey all the details in an image as we humans do. Meanwhile, with recent language models' progress, descriptive text may be an alternative to this problem. This paper delves into the effectiveness of textual representations for image understanding in the specific context of VQA. We propose to take description-question pairs as input, instead of deep visual features, and fed them into a language-only Transformer model, simplifying the process and the computational cost. We also experiment with data augmentation techniques to increase the diversity in the training set and avoid learning statistical bias. Extensive evaluations have shown that textual representations require only about a hundred words to compete with deep visual features on both VQA 2.0 and VQA-CP v2. 
% Notably, we have found that we only need a hundred words to compete with deep visual features as image representation.  
\end{abstract}


\begin{CCSXML}
<ccs2012>
   <concept>
       <concept_id>10010147.10010178.10010179</concept_id>
       <concept_desc>Computing methodologies~Natural language processing</concept_desc>
       <concept_significance>300</concept_significance>
       </concept>
   <concept>
       <concept_id>10010147.10010178.10010224.10010240.10010241</concept_id>
       <concept_desc>Computing methodologies~Image representations</concept_desc>
       <concept_significance>300</concept_significance>
       </concept>
   <concept>
       <concept_id>10010147.10010178.10010224</concept_id>
       <concept_desc>Computing methodologies~Computer vision</concept_desc>
       <concept_significance>300</concept_significance>
       </concept>
 </ccs2012>
\end{CCSXML}

\ccsdesc[300]{Computing methodologies~Computer vision}
\ccsdesc[300]{Computing methodologies~Natural language processing}
\ccsdesc[300]{Computing methodologies~Image representations}

\keywords{}


%% 6/25 %%%%%%%%%%%%%%%%%%%%%%%%%
\pdfstringdefDisableCommands{%
  \def\\{}%
  \def\texttt#1{<#1>}%
}
%%%%%%%%%%%%%%%%%%%%%%%%%%%%%%%%%%

\maketitle

\section{Introduction}

\begin{figure}[t]
  \centering
  \includegraphics[clip, width=0.95\columnwidth]{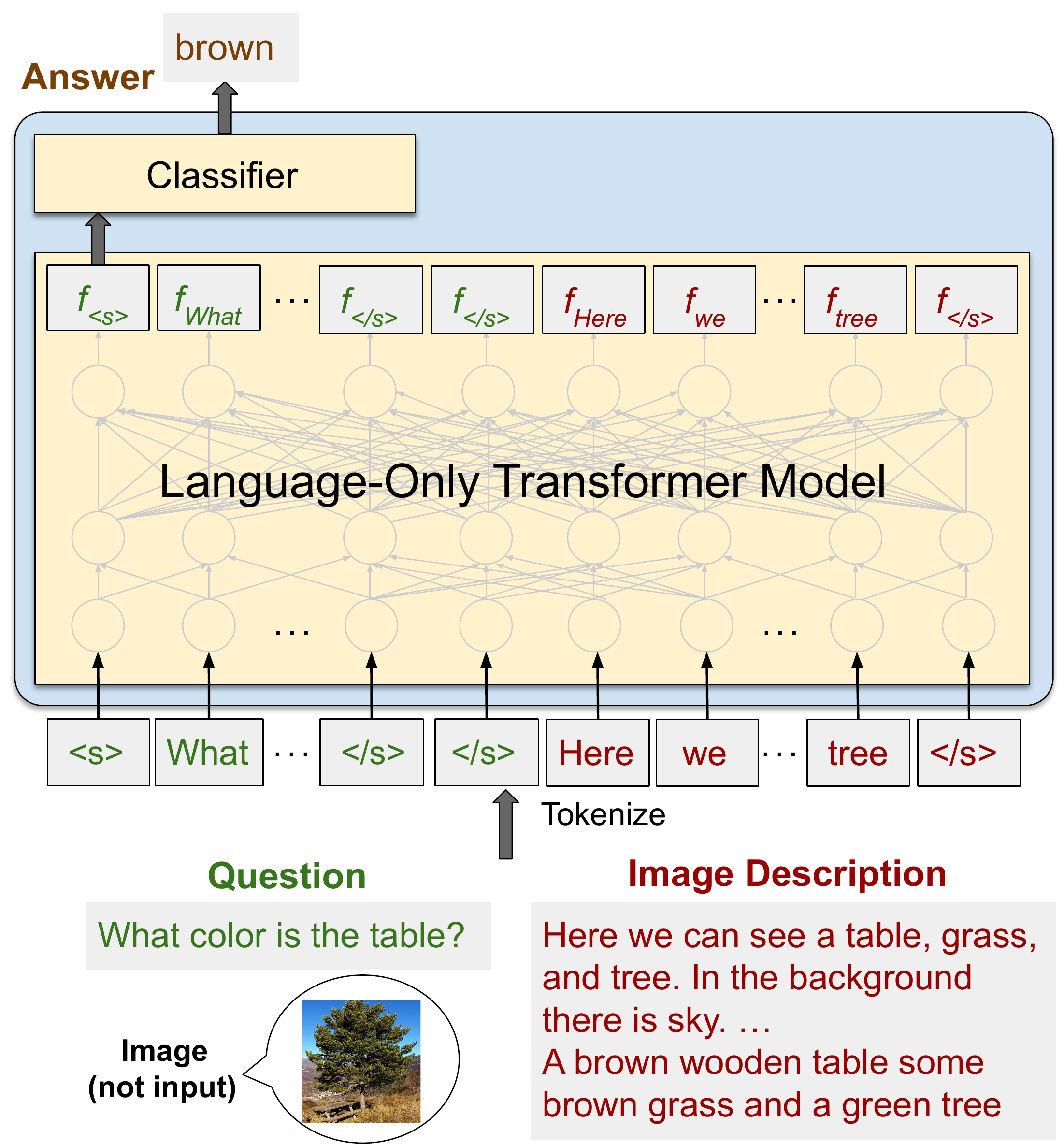} 
  \caption{Model overview.\; Our language-only model takes a question and a description as the input of a language-only Transformer model and predicts an answer accordingly. }
  \Description{Model Overview}
  \label{fig:model}
\end{figure}

The English proverb \textit{A picture is worth a thousand words} is used to convey the idea that, for humans, the meaning of a complex verbal description is often better understood with a single static image. However, identifying, understanding, and reasoning over the concepts contained in an image is still highly challenging for computer vision systems. Whereas standard deep visual features based on vectors may present some limitations to capture the rich semantic content from a picture \cite{vqacp, agrawal_analyzing, vmatter}, descriptive representations based on text may be a good alternative, especially with the recent emergence of powerful language models. This paper aims to provide a deep analysis of textual representations for image understanding tasks, and compare them with existing deep visual features techniques.

A practical task to evaluate image understanding is visual question answering (VQA). 
VQA aims to answer questions about an image's visual content, requiring a machine to understand both the question and the image.
For describing the visual content, the way in which images are represented is essential, as it highly impacts the performance of the downstream task.
Due to the bottom-up attention's success \cite{updn, ban, pythia}, deep visual features extracted by object recognition models \cite{ren2015faster} have been used as the de-facto standard for representing visual content. 
%Although models using deep visual features  have dominated the VQA leaderboard, it has been shown that they tend to learn superficial linguistic correlations , e.g., a model learns to answer {\itshape ``2''} for {\itshape `` how many X''} types of questions regardless of {\itshape ``X.''}

On the other hand, the recent progress of Transformer language models \cite{transformer,bert,xlnet, roberta} produced outstanding advances in several natural language processing (NLP) tasks, e.g., question answering \cite{SQuAD} or sentiment analysis \cite{maaslearning}. 
These language models can simultaneously represent inter-relationships between two consecutive sentences and intra-relationships between the individual words in a sentence.
With the development of Transformer language models, research on vision-and-language tasks  \cite{lu2019vilbert, lxmert, chen2020uniter, su2020vl-bert, lu2019visualbert} has shifted to explore pre-training methods to learn cross-modal representations on image-text pairs.
%They use image-text pairs for the input of Transformer language models for the models to learn image-text semantic alignments.
These pre-trained Transformers are fine-tuned on the downstream vision-and-language tasks, outperforming standard methods based on convolutional neural networks.
Within this context, one interesting question arises here: according to the popular proverb, humans may need a thousand words to describe an image; but how many words are necessary for Transformer language models?
%However, those models leverage deep visual features as the image representation, and the application of textual representations of images to VQA is still underexplored.

In this paper, we explore the effectiveness of textual representations of images and explore if they are competitive with current deep visual features.
Specifically, we conduct VQA experiments by representing images using text, instead of deep visual features.
Thus, we change the input of a VQA model from image-question pairs to image description-question pairs.
We use a RoBERTa \cite{roberta}, a state-of-the-art Transformer language-only model,  as a our VQA model. Then, the input description-question pairs are jointly fed into the model to predict an answer (Figure \ref{fig:model}). 
Besides, with the success of data augmentation methods on both VQA  \cite{mutant, chencss} and NLP \cite{eda, backtrans, qanet} tasks, we investigate the use of synthetic samples on language-only representations. As the aim of the study is to explore the viability of language-only representations in VQA, we rely on already annotated descriptions from two standard datasets \cite{cococap,localized}. Automatically generating the image descriptions, although a necessary future step, is out of the scope of this paper, and would require a whole paper on its own. We will carefully study image description generation in our future research.

Extensive experiments and analysis, including qualitative and quantitative results, have shown the effectiveness of textual representation of images. 
We have found that well-described descriptions can outperform deep visual features on two standard VQA datasets, VQA 2.0 and VQA-CP.
Additionally, we validated that it does not take a thousand words to describe an image, but about a hundred.

The key contributions of our work are summarized as follows.
\begin{itemize}[noitemsep,topsep=0pt]
\item Different from previous VQA models, we use text as image representation. 
Our setup allows us to study how well textual representation works in VQA.
We show that textual representations with about a hundred words are competitive with state-of-the-art deep visual features.
This observation promotes more interpretable representations of images in that it gives visibility to what humans easily understand.

\item To increase the diversity on the training set, we explore multiple data augmentation techniques. We propose single-modality adaptations to novel VQA data augmentation techniques, as well as we apply standard existing techniques on the NLP community.
We find that creating synthetic data consistently improves our language-only model performance. 
Prominently, back translation for questions boosts the performance by a large margin, not only in our language-only setting but also for deep visual feature-based models.

\end{itemize}

%\vspace{-10pt}
\section{Related work}

\textbf{Image Representations for VQA.}
In vision and language tasks, visual features have played a key role in leveraging the visual content of images. 
Most VQA methods \cite{stacked, updn, ban, pythia, murel, rubi, lmh} use deep visual features extracted by object detectors, such as Faster R-CNN \cite{ren2015faster}. 
These deep visual features are a set of pooled convolutional feature vectors, in which each vector represents a prominent image region.
Also, some models that use deep visual features extracted by CNN models, such as Resnet \cite{resnet}, can achieve impressive results \cite{grid, soho}.
However, VQA models using deep visual features tend to learn superficial linguistic correlations \cite{vqacp, agrawal_analyzing, vmatter}.

Building on top of the recent progress in language models, some work has also adapted Transformer models to fuse visual and textual information. 
Recent studies \cite{lu2019vilbert, lxmert, chen2020uniter, su2020vl-bert, lu2019visualbert} achieve high performance on vision-and-language tasks, such as VQA, by pre-training multi-layer Transformers with image-caption pairs. 
As a result, these models can learn general cross-modal representations.
Even though they leverage captions of images for pre-training, they rely on deep visual features for image representation. 

Similarly, image captioning \cite{youcaption, yaocaption} can be seen as a related task to text representation for images. 
Yet, converting images into text is different from our purpose of using text for representing images on VQA.
Recent work \cite{videoqa} has shown that textual representations for video question answering could outperform models based on deep visual features. 
Based on these results, we present a rigorous analysis on textual representations for VQA.

\noindent
\textbf{Data Augmentation for VQA.}
Recent work \cite{chencss, mutant} proposed data augmentation techniques for VQA to deal with language bias \cite{vqacp, agrawal_analyzing, vmatter}. 
They generated counterfactual training samples by masking, removing, or changing critical parts of the input. 
Training with these synthesized samples makes models focus on the crucial parts in an image and question, leading to alleviate the problem.
Inspired by these studies, we propose data augmentation techniques for textual representations.

% \vspace{7pt}
\noindent
\textbf{Data Augmentation for NLP.}
Various techniques of data augmentation have been proposed to improve NLP tasks.
A well-known method is back translation \cite{backtrans, qanet, fairseq} which generates new data by translating sentences to another language and back into the original language. 
Other techniques such as EDA \cite{eda} boosted the performance on text classification tasks \cite{sst2, cr, subj, trec, pc}.
In addition to the data augmentation techniques for VQA, we adopt these methods for our language-only representations, which enables us to check if data augmentation for NLP is also effective for VQA.

\section{Approach}

\begin{figure*}[t]
  \centering
  \includegraphics[clip,  width=\textwidth]{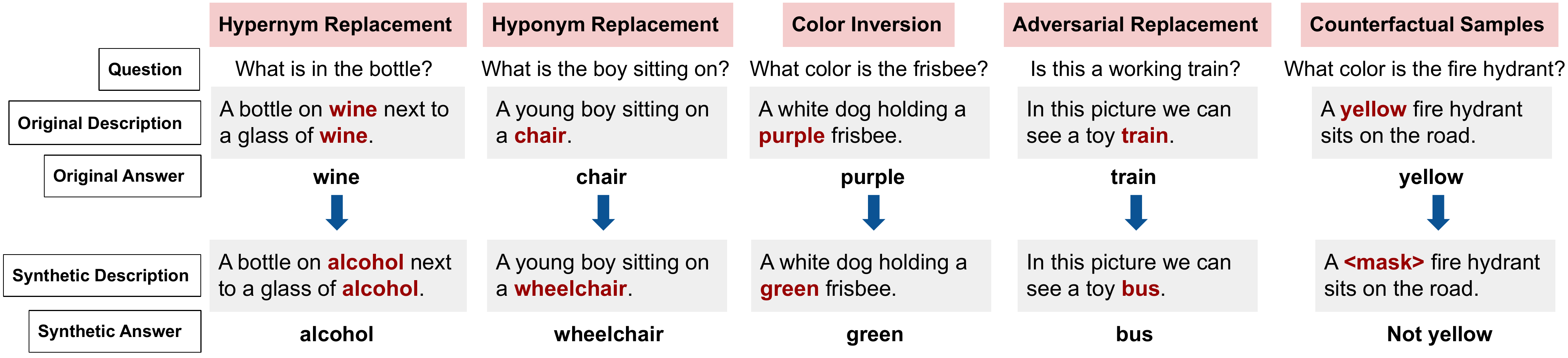}
  \caption{Generated synthetic samples using our proposed data augmentation for VQA techniques.}
  \Description{Data augmentation examples}
  \label{fig:davqa}
\end{figure*}

We introduce the text-based model with which we investigate the potential of language-only representations for VQA. An overview is shown in Figure \ref{fig:model}. 
The input, consisting of a question and a detailed description of the image, is encoded through a Transformer language-only model with several self-attention layers. 
Then, the output of the Transformer is fed into a classifier to predict an answer. We additionally propose the use of data augmentation techniques to increase the size and diversity of the training set.

\subsection{Language-Only Data}
\label{inputdata}
% \subsection{Datasets} \label{datasets}
The data for our language-only VQA framework consists of: (1) questions and answers from standard VQA datasets, (2) image descriptions representing image content, and (3) synthetic data obtained from data augmentation techniques.

%\vspace{7pt} 
\noindent
\textbf{Questions and Answers.} For the questions and answers, we use VQA-CP \cite{vqacp} and VQA 2.0 \cite{antol2015vqa} datasets. 
VQA 2.0 contains $1.1$M question-answer pairs of about $204$K images, whereas VQA-CP consists of $603$K question-answer pairs of about $219$K images. All images of VQA 2.0 and VQA-CP are from COCO dataset \cite{coco}. 
%Images are solely used to obtain the corresponding descriptions, and not input into the language-only Transformer model.
Whereas VQA 2.0 has been the de-facto standard for natural image VQA, it has also been shown to contain strong statistical biases on its training distribution \cite{vqacp, agrawal_analyzing, vmatter}, by which models obtain high accuracy by only using the first few words of a question. VQA-CP addresses this problem by re-organizing the training and validation splits to have a different answer distribution.

%\vspace{7pt}
\noindent
\textbf{Image Descriptions} % \label{image description}
We obtain image descriptions from two different corpus: COCO captions \cite{cococap} and Localized Narratives \cite{localized}. 
COCO captions contains five short captions for each image in COCO dataset~\cite{coco}, with an average length of $10.5$ words per caption. 
Captions are obtained by asking annotators to describe the important parts of the scene, without mentioning unimportant details.
Localized Narratives provides multimodal image annotations for multiple image datasets (COCO, Flickr30k \cite{flickr}, ADE20K \cite{ade}, and Open Images \cite{openimage}). The narratives are generated by asking annotators to describe an image out loud while simultaneously hovering the mouse over the region they are describing, representing the entire image,  including minor objects as opposed to COCO captions. This results in an average of $42.9$ words per narrative.
We only use narratives corresponding to COCO images.

\noindent
\textbf{Synthetic Data}
Additionally, we generate synthetic samples using data augmentation techniques to increase the size and diversity of the training data in our language-only setting. We explore multiple techniques, which can be grouped into two main categories: Data Augmentation for VQA (Section \ref{mutant-base}) and Data Augmentation for Language (Section \ref{nlp}).  
In particular, Data Augmentation for VQA (DAV) are techniques based on the latest multimodal data augmentation methods \cite{mutant,chencss} which propose to generate new images and new questions to increase the diversity in the dataset. Since we do not use the images directly, we propose single modality methods that generate synthetic descriptions that help mitigate the effects of language bias. As for Data Augmentation for Language (DAL), we leverage various established methods within the NLP community for improving language-only tasks \cite{eda, backtrans, qanet, fairseq}.

%\vspace{-10pt}
\subsection{Data Augmentation for VQA} \label{mutant-base}
We adapt data augmentation techniques for VQA \cite{mutant,chencss} to our language-only input setting. 
The aim is to generate new samples that force the model to react to essential parts in the input and provide an answer without relying on the language bias.
While image data augmentation methods for VQA \cite{teneyeccv, causalvqa, chencss, mutant} mask objects or employ GANs \cite{gan} to change the scene in an image, which is computationally expensive, we only require word replacement. 

Let $s$ denote a training triplet, i.e., ${s = (\vec{q}, \vec{d}, A)}$. $\vec{q} = [q_1, \cdots, q_Q]$ is a sequence input question with $Q$ tokens and associated with a question type $t$, 
${\vec{d} = [d_1, \cdots, d_D]}$ is a sequence input description with $D$ tokens, and $A = \{a_1, \cdots, a_N\}$ is the set of $N$ ground truth answers, where $N$ is at most 10 for the VQA 2.0 and VQA-CP datasets and can vary for different samples. We propose four techniques: (1) hypernym and hyponym replacement, (2) color inversion, (3) adversarial replacement, and (4) counterfactual samples. An example of each technique is shown in Figure \ref{fig:davqa}.

%\vspace{7pt}
\noindent
\textbf{Hypernym and Hyponym Replacement.} Following \cite{mutant}, we use hypernym and hyponym replacement to introduce similar yet semantically distinct mutations into the image descriptions.
For a given word, an hypernym covers a wider range of concepts that the original word implies, e.g.~\textit{food} is an hypernym of \textit{fruit}.
On the other hand, an hyponym covers a narrower range of meanings, e.g.~\textit{apple} is an hyponym of \textit{fruit}.

To generate new samples, %for each ground truth answer $a_j \in A$ contained in the input description $\vec{d}$ as $d_i = a_j$, we replace $a_j$ with its hypernym $h^{a_j}_e$ (or hyponym $h^{a_j}_o$). The new training triplet $s_h$ is: %
let $A_\vec{d}$ denote the set of ground truth answers $a$ such that $a$ appears in $\vec{d}$. We replace $a \in A_\vec{d}$ with its hypernym $h_e(a)$ (or hyponym $h_o(a)$) in both $A$ and $\vec{d}$. The new training triplet for hypernym replacement $s_\text{h}$ is:
\begin{equation}
s_\text{h} = (\vec{q}, \vec{d}_\text{h}, A_\text{h})
\end{equation} %
\begin{equation}
A_\text{h} = A \setminus A_\vec{d} \cup \{h_e(a)\}_{a \in A_\vec{d}}
\end{equation} %
\begin{equation}
\vec{d}_\text{h} = [d_1, \cdots, d_L] \text{ with $d_i = h_e(a)$ if $d_i = a$ for all $a \in A_\vec{d}$},
\end{equation} %
and equivalently with $h_o(a)$ for hyponym replacement. 
To avoid duplicates, we do not generate $s_\text{h}$ if $h_e(a) \in A$ (or $h_o(a) \in A$). To identify hypernyms and hyponyms, we use WordNet \cite{wordnet}.

%\vspace{7pt}
\noindent
\textbf{Color Inversion.}
For color inversion \cite{mutant}, we substitute a color word in a description with another color word. First, we manually create a set $C = \{c_1, \cdots, c_K\}$ of $K$ color words, and a set $T_C$ of question types related to colors.\footnote{$T_C = \{\text{``what color''}, \text{``what color are the''}, \text{``what color is''}, \text{``what color is the''}\}$.}
For a given training triplet $s$ and its question type $t$, if 
%a training question $\vec{q}$ is related to color, i.e., 
$t \in T_C$ and a ground truth color answer $a \in A \cap C$ is found in the input description (i.e.~$d_i = a$ for some $i$), we replace the description and the answer with a different random color word $c \neq a$. The new training triplet $s_\text{c}$ is: %
\begin{equation}
s_\text{c} = (\vec{q}, \vec{d}_\text{c}, A_\text{c})
\end{equation} %
\begin{equation}
A_\text{c} = A \setminus \{a\} \cup \{c\}
\end{equation} %
\begin{equation}
\vec{d}_\text{c} = [d_1, \cdots, d_L] \text{ with $d_i = c$ if $d_i = a$}
\end{equation} %
% \begin{equation}
% \vec{q}_c = \vec{q}
% \end{equation}
%

%\vspace{7pt}
\noindent
\textbf{Adversarial Replacement.}
For Yes/No samples, i.e. $s = (\vec{q}, \vec{d}, A)$ such that $\{\text{yes},\text{no}\} \cup A \notin \emptyset$, we replace object words $o \in O$ in description $\vec{d}$ with \textit{adversarial words}, where $O$ is the set of 80 object classes in COCO dataset~\cite{coco}. Following \cite{mutant}, we define adversarial word, $w_{\text{adv}}(o)$, as the word that is the most similar yet with a different meaning to $o$.
If $o$ (or its synonyms) are in $\vec{q}$, we change the answer from \textit{yes} to \textit{no}; otherwise the answer is not changed. Formally, the new training triplet $s_\text{a}$ is:
\begin{equation}
s_\text{a} = (\vec{q}, \vec{d}_\text{a}, A_\text{a})
\end{equation} %
\begin{equation}
A_\text{a} = 
    \begin{cases}
      ~\{\text{no}\} & \text{if $o$ is in $\vec{q}$}\\
      ~A & \text{otherwise}
    \end{cases} 
\end{equation} %
\begin{equation}
\vec{d}_\text{a} = [d_1, \cdots, d_L] \text{ with $d_i = w_{\text{adv}}(o)$ if $d_i = o$},
\end{equation} %
% \begin{equation}
% \vec{q}_a = \vec{q}
% \end{equation}
%
where $w_{\text{adv}}(o)$ is selected as the closest word to $o \in O$ according to the Euclidean distance between their Glove embeddings \cite{glove}.

Differently from \cite{mutant}, we do not generate adversarial samples from questions, as changing or masking a word in a question leads to a new answer, which may not be determined automatically, e.g., \textit{``How many bins?''} \textrightarrow \textit{``How many pens?''}.

\noindent
\textbf{Counterfactual Samples.} % \label{css-base}
Counterfactual samples \cite{chencss} are modifications of questions or (part of) images that make the original question-answer pairs irrelevant. 
We generate counterfactual training samples by adapting \cite{chencss} to language-only description-question pairs. Given the training triplet $s = (\vec{q}, \vec{d}, A)$, we generate counterfactual samples $s_{\text{css}_q}$ and $s_{\text{css}_d}$ from the query and description:
\begin{align}
s_{\text{css}_\text{q}} = (\vec{q}_{\text{css}_\text{q}},  \vec{d}, A_{\text{css}_\text{q}}) \\
%\end{align} %
%
%\begin{equation}
s_{\text{css}_\text{d}} = (\vec{q}, \vec{d}_{\text{css}_\text{d}}, A_{\text{css}_\text{d}})
\end{align} %

To find $s_{\text{css}_\text{q}}$, we feed $\vec{q}$ and $\vec{d}$ into a trained Transformer VQA model on the target dataset, $\mathcal{M}$, and obtain the contribution of each word in $\vec{q}$ to the answer set $A$ with Grad-CAM \cite{gradcam}. The top-D words with highest contribution are selected as a set $\Omega_\text{q}$ of critical words. Two new questions $\vec{q}_{\text{css}_\text{q}}^+$ and $\vec{q}_{\text{css}_\text{q}}^-$ are generated by masking all the words in $\vec{q}$ not contained in  $\Omega_\text{q}$, and masking all the words contained in $\Omega_\text{q}$, respectively:
\begin{align}
\vec{q}_{\text{css}_\text{q}}^+ = [q_1, \cdots, q_L] &\quad \text{ with } q_i = \text{<mask>} \quad \text{for all } q_i \notin \Omega_\text{q} \label{eq:css_plus}\\
%\end{equation} %
%
%\begin{equation}
\vec{q}_{\text{css}_\text{q}}^- = [q_1, \cdots, q_L] &\quad \text{ with } q_i = \text{<mask>} \quad \text{for all } q_i \in \Omega_\text{q},
\end{align} %
%\begin{equation}
%\vec{q}_{\text{css}_q} = \vec{q}_{\text{css}_q}^-
%\end{equation} %
%
where $\vec{q}_{\text{css}_q}^-$ is used as this CSS sample's question $\vec{q}_{\text{css}_q}$ and in Eq.~(\ref{eq:css_plus}), the first few words that correspond to the question type (e.g., \textit{what color is}) are not masked, as in \cite{chencss}. 

To find $A_{\text{css}_\text{q}}$, $\vec{q}_{\text{css}_\text{q}}^+$ and $\vec{d}$ are again fed into $\mathcal{M}$ to obtain the score for each candidate answer. The top-J scoring answers are removed from the original set of answers. Specifically, letting $\mathcal{M}_J(\vec{q}_{\text{css}_\text{q}}^+, \vec{d})$ be the set of the top-J answers, the new ground truth answers is given by 
\begin{equation}
A_{\text{css}_\text{q}} = A \setminus \mathcal{M}_J(\vec{q}_{\text{css}_\text{q}}^+, \vec{d}).
\end{equation} %
For $s_{\text{css}_\text{d}}$, the values of $\Omega_\text{d}$, $\vec{d}^+_{\text{css}_\text{d}}$, $\vec{d}^-_{\text{css}_\text{d}}$, and $M_N(\vec{q}, \vec{d}^+_{\text{css}_\text{d}})$ are found in the same way.

\subsection{Data Augmentation for Language} \label{nlp}
Given that the input to our VQA model is solely based on the language modality, we also explore NLP data augmentation techniques.
Among all existing techniques, we adopt three of the most popular and successful ones \cite{eda, varunaug, qizhebt, qanet, contextual, deepcontext}: (1) EDA, (2) back translation, and (3) word replacement/insertion via contextual word embedding.
Each technique is applied to either the description or the question of the input triplet $s$ to generate new samples:
\begin{align}
s_{\text{nlp}_\text{q}} = (\vec{q}_{\text{nlp}}, \vec{d}, A) \\
%\end{align} %
%
%\begin{equation}
s_{\text{nlp}_\text{d}} = (\vec{q}, \vec{d}_{\text{nlp}}, A)
\end{align} %
where $\vec{q}_{\text{nlp}}$ and $\vec{d}_{\text{nlp}}$ are the question and the description, respectively, after applying one of the transformations below. Examples of synthetic samples are shown in Table \ref{tab:nlpsamples}.

\begin{table}
  \caption{Examples of Data Augmentation for Language.\; }
  \label{tab:nlpsamples}
  \begin{tabular}{ll}
    \bhline{0.7pt}
    DAL Method & Synthetic Question \\
    \bhline{0.7pt}
    Original & What is the giraffe standing behind?\\
    \midrule
    EDA (SR) &  What is the camelopard standing behind?\\ 
    EDA (RI) & What is the giraffe abide standing behind? \\
    EDA (RS) & What is the standing giraffe behind?\\
    EDA (RD) & What is the standing behind?\\
    BT & What's behind the giraffe?\\
    CWR & What is the giraffe tree behind?\\
    CWI & What is the giraffe standing silently behind?\\
    \bhline{0.7pt}
  \end{tabular}
  \vspace{-5pt}
\end{table}

\vspace{7pt}
\noindent
\textbf{EDA} (Easy Data Augmentation) \cite{eda} is composed of 4 operations:
\begin{itemize}[noitemsep,topsep=0pt]
    \item \textit{Synonym Replacement} randomly chooses $n$ words from the input sentence and replaces them with their synonyms.
    \item \textit{Random Insertion} finds a random synonym of a random word in the sentence and inserts it in an arbitrary position.
    \item \textit{Random Swap} chooses two words and swaps them.
    \item \textit{Random Deletion} removes words with probability $p$.
\end{itemize}
While EDA consists of simple word replacement-based techniques, it has been shown to improve text classification performance in low-resource tasks \cite{eda}.
%We generate a new description or question by applying one of these four operations randomly.
For each sample, we apply one of these four operations randomly.

%\vspace{7pt}
\noindent
\textbf{Back Translation} \cite{backtrans, qanet} translates a sentence into another language and then translates it back into the original language.
As shown in \cite{qizhebt, qanet}, back translation can generate diverse paraphrases while preserving the original sentences' semantics, resulting in significant performance improvements in various NLP tasks \cite{SQuAD,andrewtext, xiangtext}.
For the implementation of back translation, we use the python nlpaug library \cite{nlpaug} based on fairseq \cite{fairseq, wmt}.
The translator is based on the big Transformer architecture \cite{transformer} and translates a given sentence to German and back into English.
If the sentence has not changed after applying the back translation, we discard it.

%\vspace{7pt}
\noindent
\textbf{Contextual Word Replacement/Insertion} replaces or inserts words by considering the context. 
To obtain the words that match the context, it uses deep bidirectional language models \cite{bert, deepcontext, xlnet}.
We again use the implementation of contextual word embedding provided by python nlpaug library \cite{nlpaug}.
We replace or insert the most similar words of randomly selected words in the description or question to generate a new one,
where we use pre-trained XLNET \cite{xlnet} in the library to obtain the contextual word embeddings.

\subsection{Language-Only VQA Model} \label{transformer}
As opposed to most VQA models, which take image-question pairs as input, the input of our language-only VQA model is description-question pairs. 
Given the original VQA triplet $(\vec{q}, I, A)$, where $I$ is the image associated with the question, we first create the image description $\vec{d}$ by concatenating the narrative of Localized Narratives and the captions of COCO captions.
The question and the image description are compiled into a single sequence, $\vec{l}$, by inserting a classifier token {\small\texttt{<s>}} at the beginning of the sequence, and end of sentence tokens {\small\texttt{</s>}} as:
\begin{equation}
\vec{l} = {\small\texttt{<s>}} + \vec{q} + {\small\texttt{</s>}} + {\small\texttt{</s>}} + \vec{d} + {\small\texttt{</s>}}
\end{equation}
where ``$+$'' indicates concatenation. The input sequence is fed into our Transformer-based language model $\mathcal{T}$ to obtain sequence $\vec{f}$ of embeddings, i.e.,
\begin{equation}
\vec{f} = \mathcal{T}(\vec{l}). 
\end{equation}
Then, the embedding corresponding to the classifier token {\small\texttt{<s>}} is fed into classifier $\mathcal{C}$, to make a prediction:
\begin{equation}
\boldsymbol{\rho} = \mathcal{C}(\vec{f}_{\small\texttt{<s>}})
\end{equation}%
Unless otherwise stated, we use large $\text{RoBERTa}$ \cite{roberta} as our Transformer language model, which has been shown to be one of the best performing models on NLP tasks \cite{comptf_cr, comptf_tc}.

\section{Experiments}
We conducted five main experiments to validate the effectiveness of textual representations for VQA: (1) comparison between different image descriptions, (2) comparison between data augmentation techniques, (3) comparing textual representations with deep visual features, (4) evaluating the impact of data augmentation for questions on deep visual feature-based models, and (5) comparison between language-only Transformers.

For RoBERTa hyperparameters, we followed \cite{roberta}, only changing the number of training epochs to $20$. As a classifier, we used a multi-layer perceptron with two fully-connected layers with $2048$ hidden units, and Swish activation function \cite{swish} between them. We use softmax cross entropy over the answer vocabulary for the loss function. 
The parameters for the data augmentation techniques were as follows: for counterfactual samples, we set top-D=$10$ to select critical words and top-J=$5$ for assigning new answers; for EDA, we set the rate of the words to be changed to $0.1$, following the recommended value in the original paper \cite{eda}. Unless otherwise stated, the input of our model consists of the whole sequence of question, narrative, and five captions.

Results are presented in terms of accuracy, following the standard protocol \cite{antol2015vqa}.
For the datasets split, the VQA-CP v2 training set contains $438$K questions about $120$K images, whereas the test set has $220$K questions about $98$K images. As for VQA 2.0, the training set includes $443$K questions about $82$K images, while the val set contains $214$K questions about $40$K images. As the VQA 2.0 test set requires access to external server, we report results on the val set.

\subsection{Describing an Image with Words}

\begin{table}[t]
  \caption{Image Description Evaluation. Length denotes the mean number of tokens in the image descriptions.}
  \label{tab:langinputeval}
  \begin{tabular}{lrr}
    \bhline{0.7pt}
    Image Description & Length & Accuracy\\
    \bhline{0.7pt}
    None {\small(Question-Only)} & - & 21.39\\
    % \midrule
    1 Caption & 10.5 & 35.31\\
    2 Captions & 21.0 & 38.49\\
    3 Captions & 31.5 & 40.09\\
    4 Captions & 42.0 & 41.93\\
    5 Captions & 52.5 & 42.34\\
    Narrative &42.9 & 36.45\\
   % LN + 1 Caption & 58.2 & \\
    Whole {\small(Narrative + 5 Captions)} & 95.3 & \textbf{43.64} \\ 
    \bhline{0.7pt}
  \end{tabular}
\end{table}

\begin{figure}[t]
  \centering
  \includegraphics[clip, scale=0.5]{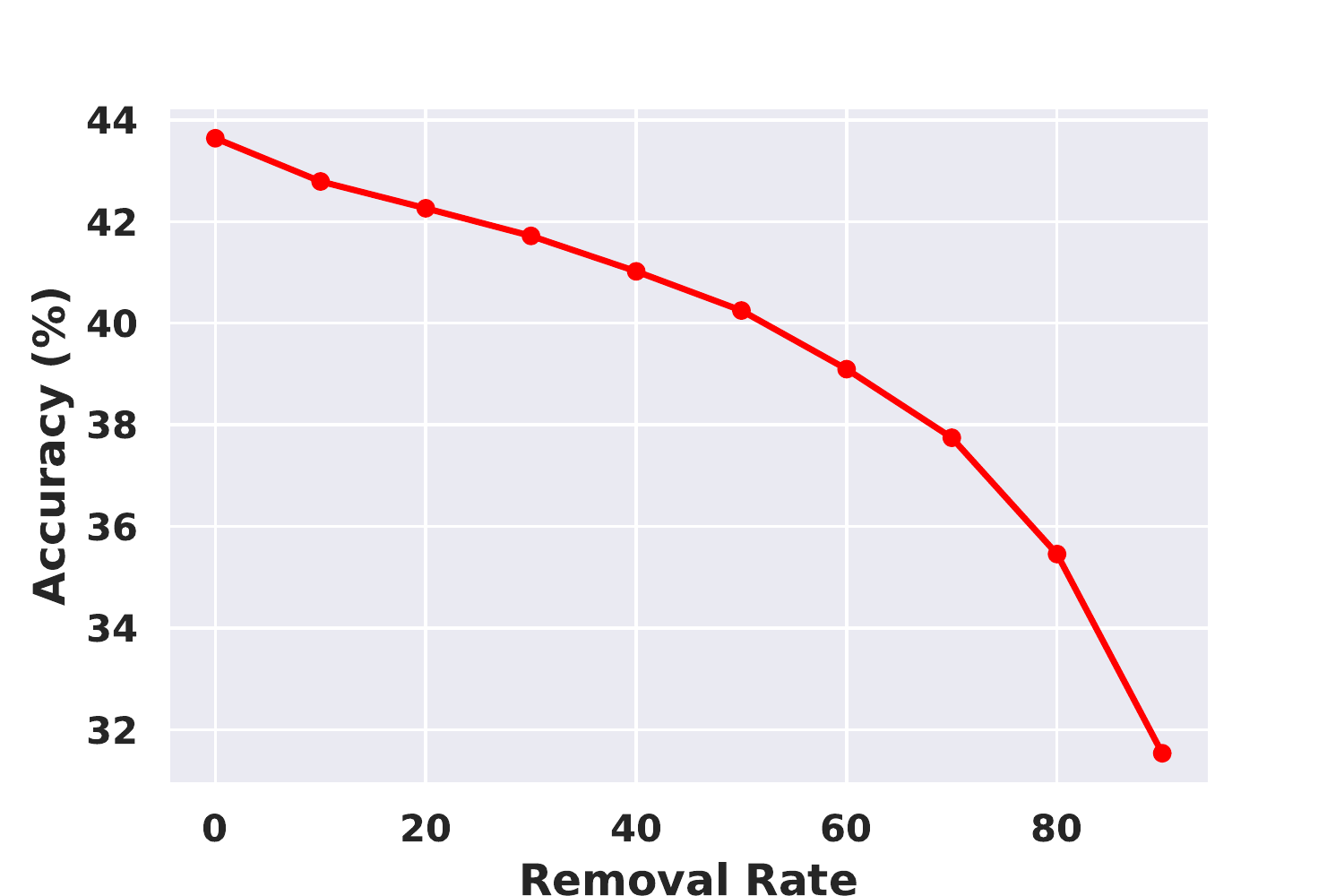}
  \caption{Effect of truncating the input sequence on the VQA CP v2 test set. Removing words from the input has a negative effect on the accuracy, suggesting that every word in the sequence contributes to the overall accuracy.}
  \Description{Answer overlap}
  \label{fig:ablation}
\end{figure}

% \textbf{Settings.}
Here, we investigated the relationship between the quality of the image descriptions and the accuracy on the VQA task by conducting two experiments: (1) comparing the performance of different types of image descriptions and (2) studying the correlation between accuracy and the length of the input descriptions.

%\vspace{7pt}
\noindent
\textbf{Image descriptions.}
We first evaluate the performance of different language-only inputs. Specifically, we consider the following inputs: only the question, question and 1 to 5 random captions, question and narrative, and the whole input (question, narrative, and 5 captions). Results on the VQA-CP v2 test set are reported in Table~\ref{tab:langinputeval}, along with the average sequence length.

The whole input, consisting on merging the narrative with the 5 captions with an average of $95.3$ tokens per sample, performs the best, which indicates that the two datasets contain complimentary useful information for VQA.
When comparing captions and narratives, we find that the former produce better results with fewer words. Specifically, by just using two captions, the performance is already better than with narratives, even though the average number of words is about half. This confirms that the VQA dataset contains a substantial amount of questions about the general content of the image, rather than its specific details, as the captions in COCO dataset, in contrast to narratives, focus mostly on the prominent areas in the scene. When using a single caption, the accuracy improves by $13.92$\% compared to the Question-Only input, showing that a short description of about $10$ words already contains enough information to answer correctly one in three questions.

%\vspace{7pt}
\noindent
\textbf{Input length}
For studying the correlation between accuracy and length of the input, we progressively truncate the image descriptions at test time and evaluate their performance on a model trained on the full input without truncation (question, narrative, and 5 captions). 
To truncate the input, we rearrange the sentences randomly, truncate the words at the end of the rearranged sequence, and put the sentences back in their original order to avoid losing contextual information. Questions are not truncated in any case.
Results are shown in Figure \ref{fig:ablation}, where it can be seen that the accuracy decreases as the image descriptions are trimmed.
In particular, the accuracy decreases linearly up to $60$\%. For more than $60$\% of removal rate, the model cannot deal with the information loss and the accuracy decreases rapidly. This tendency suggests that most words on the input have a positive contribution to the overall accuracy.

\begin{table*}
  \caption{Data augmentation results on the VQA-CP v2 test set.\; In DAL, D and Q denote when applied to descriptions or questions, respectively. Gap is the overall accuracy difference compared to the accuracy when not using synthetic samples. }
  \label{tab:augresult}
  \begin{tabular}{clrrrrrrr}
    \bhline{0.7pt}
    & Input Data & Num. Synthetic & Num. Total &Yes/No&Number&Other &Overall & Gap\\
    \bhline{0.7pt}
    & Narrative + 5 Captions &- &438,183 & 45.13 & 20.06 & 49.33& 43.64 &-\\
    \midrule
    \multirow{6}*{\rotatebox[origin=c]{90}{DAV}}& \hspace{2pt} w/ Hyponym Replacement & 132,570 & 570,753  & 45.65 & 25.36 & 50.52 & 45.26 & \textcolor[rgb]{0.2, 0.5, 0.2}{$+$1.62}\\
    & \hspace{2pt} w/ Hypernym Replacement& 23,869& 462,052  & 47.28 & 17.69 & 49.10 & 43.70 & \textcolor[rgb]{0.2, 0.5, 0.2}{$+$0.06}\\
    & \hspace{2pt} w/ Hyponym and Hypernym Replacement& 183,944 & 622,177 & 45.80 & 21.46 & 51.15 & 45.06 &\textcolor[rgb]{0.2, 0.5, 0.2}{$+$1.42}\\
    & \hspace{2pt} w/ Color Inversion& 19,308 & 457,491  &45.61 &19.93 &50.60 &44.47 & \textcolor[rgb]{0.2, 0.5, 0.2}{$+$1.06}\\
    & \hspace{2pt} w/ Adversarial Word Replacement &169,929 & 608,112 & 44.71 &19.84 &50.03 &43.93  & \textcolor[rgb]{0.2, 0.5, 0.2}{$+$0.29} \\
    & \hspace{2pt} w/ Counterfactual Samples &438,183 & 876,366 & 44.20 & 19.84 & 52.07 & 44.86 & \textcolor[rgb]{0.2, 0.5, 0.2}{$+$1.22}\\
    \midrule
    \multirow{8}*{\rotatebox[origin=c]{90}{DAL}} & \hspace{2pt} w/ EDA (D)& 438,183 & 876,366 &44.68 &20.64 &50.08 & 44.02 & \textcolor[rgb]{0.2, 0.5, 0.2}{$+$0.38}\\
    & \hspace{2pt} w/ EDA (Q)& 438,183 & 876,366  &46.86 &23.50 &50.62 &45.39 & \textcolor[rgb]{0.2, 0.5, 0.2}{$+$1.75}\\ 
    & \hspace{2pt} w/ Contextual Word Replacement (D)& 438,183 & 876,366& 44.69 & 19.40 & 48.91 &43.18 & \textcolor{red}{$-$0.46} \\
    & \hspace{2pt} w/ Contextual Word Replacement (Q)& 438,183 & 876,366 &46.09 &22.49 &49.10&44.16 & \textcolor[rgb]{0.2, 0.5, 0.2}{$+$0.52}\\
    & \hspace{2pt} w/ Contextual Word Insertion (D)&438,183 & 876,366 &45.15 &19.31 &48.86 &43.27 & \textcolor{red}{$-$0.37} \\
    & \hspace{2pt} w/ Contextual Word Insertion (Q)&438,183 & 876,366 &45.86 & 21.44&51.10 &45.05 & \textcolor[rgb]{0.2, 0.5, 0.2}{$+$1.41}\\
    & \hspace{2pt} w/ Back Translation (D)&438,183 & 876,366 & 45.28 & 21.01 & 50.89 &44.70 & \textcolor[rgb]{0.2, 0.5, 0.2}{$+$1.06}\\
    & \hspace{2pt} w/ Back Translation (Q)& 293,811 & 731,994 & \textbf{62.43} &\textbf{27.15} &\textbf{51.84} &\textbf{51.16} &\textbf{\textcolor[rgb]{0.2, 0.5, 0.2}{$+$7.52}} \\
    \bhline{0.7pt}
  \end{tabular}
\end{table*}

\subsection{Use of Synthetic Samples} \label{data_aug result}

% \textbf{Settings.}
We evaluate the performance when augmenting the VQA-CP v2 training set with synthetic samples. 
% We used the training set of VQA-CP v2 \cite{vqacp} that contains about 438K questions about $120$K images to generate samples.
The input consists on the whole description with the narrative and the five captions.

Results for each of the proposed data augmentation techniques are shown in Table~\ref{tab:augresult}.
Whereas most techniques are able to increase the accuracy with respect to the baseline (when no data augmentation is used), back translation for questions has by far the best results, with a gain of $7.52$ points on overall, and a boost of $17.30$ points on Yes/No questions. 
As back translation is the only data augmentation technique in Table~\ref{tab:augresult} that generates new samples while maintaining the original semantics, its impressive performance points us to the importance of 1) having diversity within the training questions set, whereas at the same time,  2) a correct relationship between the triplet question-description-answer semantics.

Most of the other DAL techniques, except contextual word replacement/insertion for descriptions, also increase the accuracy with respect to the baseline. An interesting observation is that when data augmentation is applied to the questions, the performance consistently improves more than when it is applied to descriptions. As for the DAV techniques, while all of them improve accuracy, hyponym replacement turns out to be the most powerful one, with an overall accuracy gain of $1.62$. Surprisingly, using hyponym and hypernym replacement together does not perform better than hyponym replacement by its own, showing that in some cases the combination of synthetic samples obtained from different techniques may be harmful.

\subsection{Comparison against Deep Visual Features}

%maybe gap is not necessary.

\begin{table*}
  \caption{Comparison of language-only representations with standard deep visual features. $^{\ast}$ indicates our re-implementations.} %Accuracies (\%) on VQA-CP v2 test set and VQA 2.0 val set, containing $214$K questions about $40$K images, of our model and state-of-the-art models using deep visual features as image representation. Models with $^{\ast}$}
  \label{tab:vqa-vqacp}
  \begin{tabular}{lcrrrrcrrrr}
    \bhline{0.7pt}
    &&\multicolumn{4}{c}{VQA-CP v2 test}&&\multicolumn{4}{c}{VQA 2.0 val}\\
    \cline{3-6}
    \cline{8-11}
    Model&&Yes/No&Number&Other&Overall&&Yes/No&Number&Other &Overall\\ 

    \bhline{0.7pt}
    HAN \cite{han} && 52.25 & 13.79 & 20.33 & 28.65 && - & - & - & - \\
    RAMEN \cite{ramen} && - & - & - & 39.21 && - & - & - & - \\
    MuRel \cite{murel}  && 42.85 & 13.17 & 45.04 & 39.54 && - & - & - & 65.14 \\
    UpDn \cite{updn} && 42.27 & 11.93 & 46.05 & 39.74 && 81.18 & 42.14 & 55.66 & 63.48  \\
    ReGAT \cite{regat}    && - & - & - & 40.42 && - & - & - & 67.18 \\
    $\text{BAN}^{\ast}$ \cite{ban} && 43.14 & 13.63 & 46.92& 40.74 && 83.19 & 48.13 & 57.52& 65.93  \\
    $\text{LXMERT}^{\ast}$ \cite{lxmert}&& 42.01 & 14.16 & 48.34 & 41.28  && 83.30 & 46.15 & 56.91& 65.31 \\
    NSM \cite{nsm}   && - & - & - & \textbf{45.80} && - & - & - & - \\
        \hline
    Ours {\small (Narrative + 5 Captions)}  && 45.13 & 20.06 & 49.33& 43.64 && 87.91 & 56.47 & 59.43& \textbf{69.74} \\
     \bhline{0.7pt}
  \end{tabular}
\end{table*}

% \textbf{Settings.}
We compare our whole language-only representations (question, narrative, and five captions) with state-of-the-art VQA models based on deep visual features on both the VQA-CP v2 and the VQA 2.0 datasets. 
For a fair comparison, we do not include the models developed to mitigate language bias \cite{murel, rubi, lmh}, as these modules can be added as a plug-in extension to any other method, including ours. For the same reason, we do not use data augmentation.

% \vspace{7pt}
% \noindent
% \textbf{Results.}
Results are reported in Table~\ref{tab:vqa-vqacp}.
Our model outperforms most of the baselines that use deep visual features on both VQA-CP v2 and VQA 2.0. For the accuracy on VQA-CP v2, NSM~\cite{nsm} performs slightly better than our language-only model. 
This result verifies that the textual representations of the images are effective for both cases where the distribution of answers per question is different between the training and test set and when it is not.
This also shows the textual representations capture well the image content, such that they are competitive with deep visual features.

\subsection{Back Translation for Other Models}

\begin{table}
\setlength{\tabcolsep}{3pt}
  \caption{Results of applying back translation to different VQA models in the VQA-CP v2 test set. Gap represents the improvement by training with the synthetic samples.}
  \label{tab:deepbt}
  \begin{tabular}{lrrrrr}
    \bhline{0.7pt}
    Model&Yes/No&Number&Other&Overall&Gap \\
    \bhline{0.7pt}
    BAN \cite{ban} & 43.14 & 13.63 & 46.92 & 40.74 & - \\
    \hspace{2pt} w/ back translation & 47.87 & 16.27 & 48.76 & 43.57 & +2.83 \\
    % \midrule
    LXMERT \cite{lxmert} & 42.01 & 14.16 & 48.34 &41.28 & - \\
    \hspace{2pt} w/ back translation &  48.81 & 16.16 & 49.75 &44.35 & +3.07 \\
    %  \midrule
    Ours & 45.13 & 20.06 & 49.33& 43.64& - \\
    \hspace{2pt} w/ back translation & 62.43 & 27.15 & 51.84& 51.16 & +7.52 \\
    \bhline{0.7pt}
  \end{tabular}
\end{table}

% \textbf{Settings.}
The results of data augmentation techniques in Section \ref{data_aug result} showed that applying back translation to questions is very effective to boost accuracy in our language-only setting. 
As back translation can be easily applicable to any other settings, we explore if its benefits are transferable to standard VQA models. We conduct experiments on BAN~\cite{ban} and LXMERT~\cite{lxmert} by adding synthetic back translated samples (applied to questions) to the training set.

% \vspace{7pt}
% \noindent
% \textbf{Results.}
Results are shown in Table~\ref{tab:deepbt}.
We can see that training BAN and LXMERT with synthesized back translated samples  improves the performance by a large margin, with gaps of $2.83$ and $3.07$ points, respectively.
All types of questions are benefited from this technique, but especially Yes/No with an improvement of about $6.80$.
This tendency is consistent with the findings in our model, strongly indicating that leveraging additional samples containing questions with the same meaning but expressed in a different way is extremely beneficial. 
In other words, adding diversity to the questions increases the model's ability to interpret the questions.

\subsection{Language Transformers}

\begin{table}
  \caption{Evaluation of different language-only Transformer models on the VQA-CP v2 test set.}
  \label{tab:tfcomp}
  \begin{tabular}{lrrrr}
    \bhline{0.7pt}
    Model&Yes/No&Number&Other&Overall \\
    \bhline{0.7pt}
    BERT base & 42.66 & 16.19 & 45.66 & 40.29 \\
    BERT large & 42.72 & 17.43 & 48.47 & 42.06 \\
    % \midrule
    XLNET base & 43.49 & 17.61 & 48.45 & 42.30 \\
    XLNET large & 44.58 & \textbf{20.67} & \textbf{50.52} & \textbf{44.23}\\
    % \midrule
    RoBERTa base & 44.39 & 17.46 & 48.74 & 42.70 \\
    RoBERTa large & \textbf{45.13} & 20.06 & 49.33& 43.64 \\
    \bhline{0.7pt}
  \end{tabular}
\end{table}

% \textbf{Settings.}
Finally, we compare the performance of prevailing language-only Transformer models: BERT \cite{bert}, XLNET\cite{xlnet}, and RoBERTa\cite{roberta}, both in their base and large versions.
All the models are exposed to the same input, consisting of the question, the narrative, and the five captions. 

Results are shown in Table~\ref{tab:tfcomp}. All the models present a similar behavior. XLNET large has the best performance, and its accuracy is higher $0.59$\% compared to RoBERTa large.
However, while the improvement is minor, the computational time of XLNET large is about $2.7$ times that RoBERTa.
Considering this, it is reasonable to use RoBERTa large to save the training time while obtaining relatively similar results.

\section{What do Textual Representations Learn?}
In this section, we analyze the features and advantages of text for representing images compared with deep visual features.
To this end, we conducted the following two analysis: (1) exploring the overlap of the predictions between our language-only model with the models using deep visual features, (2) investigating some visual examples for qualitative analysis. 
The input to our models is formed by combining the narrative with the five captions.

\begin{figure*}[t]
  \centering
  \includegraphics[clip, scale=0.365]{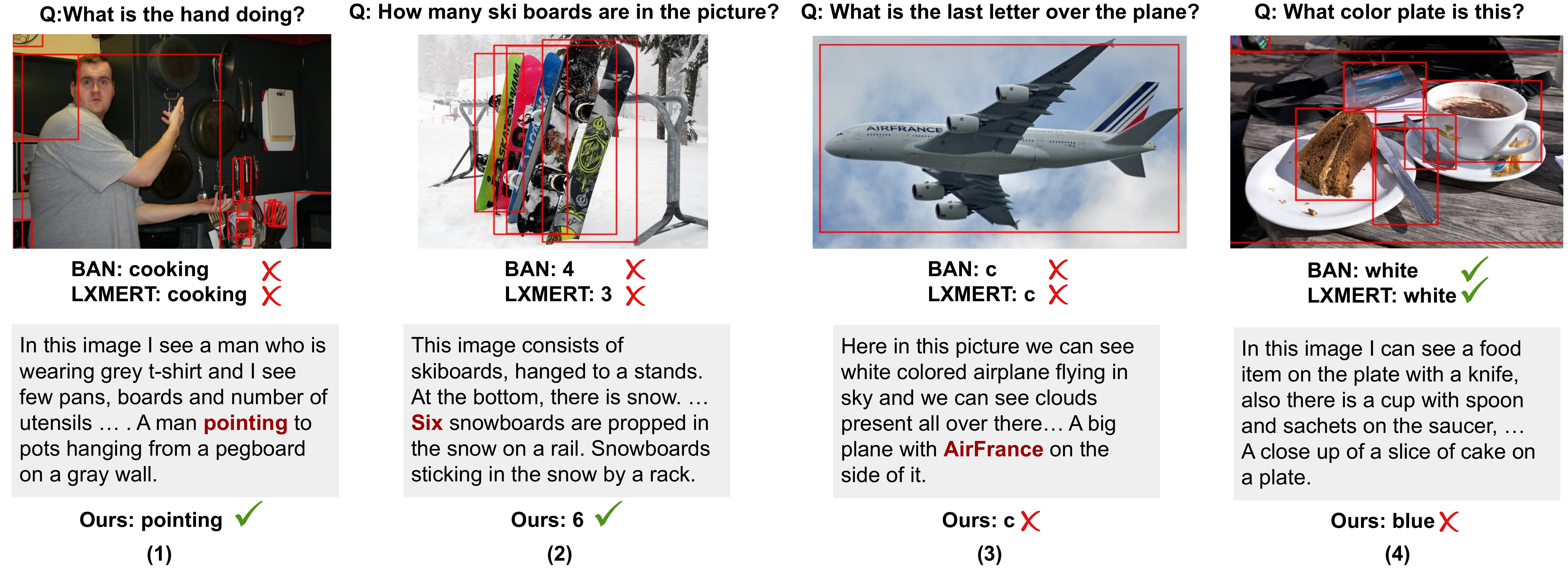}
  \caption{Qualitative Comparison.\; The red boxes in the images denote the result of detection by Faster R-CNN. Only bounding boxes with a confidence score greater than $\bm{0.5}$ are shown. The shown descriptions are extraction of the actual descriptions. The red highlighted words are the relevant words to the answers.}
  \Description{Qualitative comparison}
  \label{fig:qualitative}
%   \vspace{10pt}
\end{figure*}

\subsection{Error Analysis}

\begin{figure}[t]
  \centering
  \includegraphics[clip, width=\columnwidth]{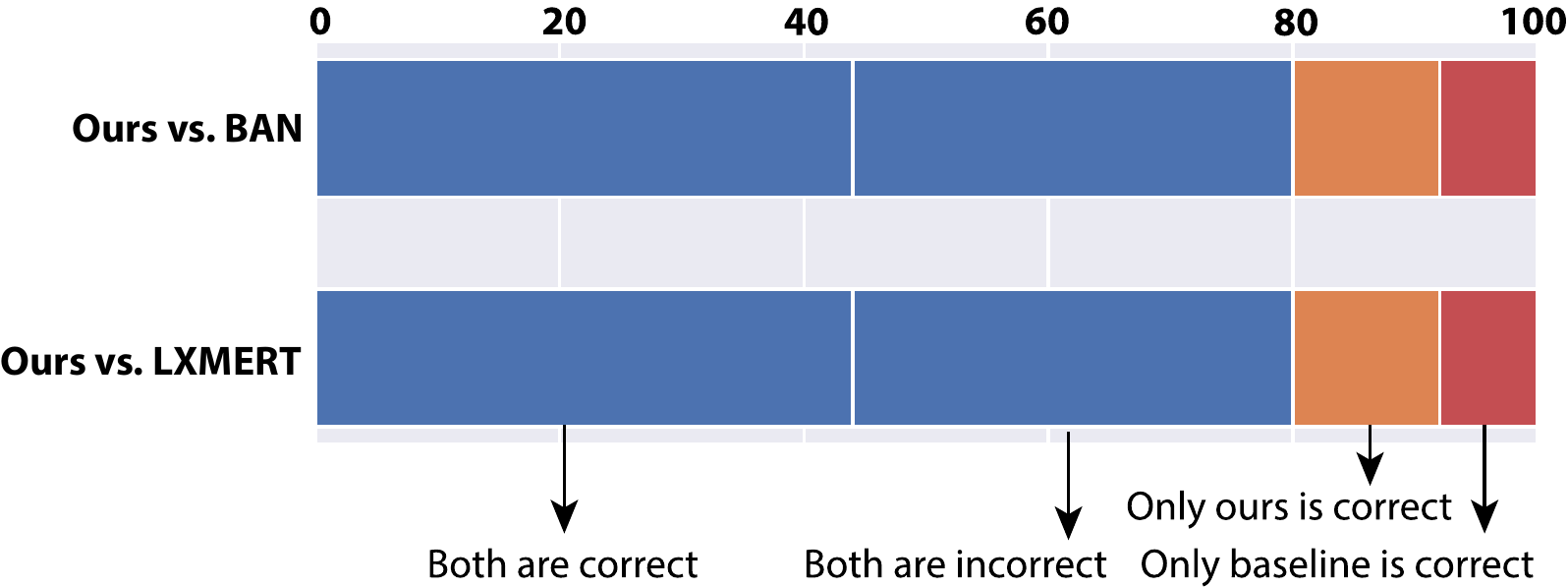} 
  \caption{Answer Overlap.\; The ratio of same or different predictions between our language-only model and BAN or LXMERT. Bar graphs denote the proportion of the total. }
  \Description{Answer overlap}
  \label{fig:overlap}
\end{figure}

We compare our model that only takes text as input with models that use deep visual features and investigate whether our model makes the same or different mistakes.  
We adopt BAN and LXMERT as representative models leveraging deep visual features.
Whereas BAN was one of the top-performing models before the emergence of the Transformer-based vision-and-language models, LXMERT is one of the state-of-the-art pre-trained models based on multi-modal Transformers extracting cross-modal features. 
LXMERT can be seen as the middle model between our language-only model and BAN, as LXMERT is pre-trained by taking a pair of image and caption as input.
%As the description of an image in our model, we use narratives for Localized Narratives and five captions for COCO.

Figure \ref{fig:overlap} compares our model with BAN and LXMERT in terms of the consistency of correct and incorrect answers.
Note that the ratios of correct answers are higher than the results in Table \ref{tab:vqa-vqacp} because the original evaluation method uses a non-binary accuracy metric for each prediction \cite{antol2015vqa}, while for this comparison, we employed a binary one where a prediction is counted to be correct if the predicted answer is in $A$ for simplicity. 
We can see that our model shows the relatively high consistency, and the ratios of consistent/inconsistent predictions are almost same for both BAN and LXMERT.
Our model and both baselines are consistent for $80$\% of the questions (blue part).
The ratio that only our model answering correctly is $12$\% for both baselines (orange part), while the ratio reduces to $8$\% (red part).
This provides further evidence that the textual representation of images can compete with deep visual features.

\subsection{Qualitative Analysis}

We show some qualitative examples in Figure \ref{fig:qualitative}.
We compare our model's predictions with the predictions of BAN and LXMERT. 
Example (1) asks what the man is doing with his hand.
The image description contains the expression of what the man is doing ( {\itshape``A man pointing to pots...''}), result in our language-only model answering correctly.
On the other hand, BAN and LXMERT fail to make the correct prediction, which they answer {\itshape``cooking''} nevertheless the man is not cooking. 
The object detector can detect the cooking tools, so the models may cause this mistake guess from the utensils, not from the man's move.
In example (2), the image description also describes the critical information to answer the question ({\itshape``Six snowboards''}), while the object detector cannot detect all objects. 
On the contrary, examples (3) and (4) show the limitations of our language-only model. 
Example (3) requires reading the letters on the side of the plane. The image description contains the necessary word ({\itshape``AirFrance''}) to answer the question but fails to make a correct prediction.
This result poses the lack of language-only Transformer models' ability to understand the contents and relations between the question and description. 
Additionally, example (4) shows the importance of the image descriptions' quality. Our model fails to correctly answer because there is no information to answer the question. 
From these observations, utilizing well-described text as image representation has the advantage against the deep visual features when answering the questions that deep visual features do not work well.
On the other hand, we identify the limitation of our language-only model for understanding the text input.

\subsection{Limitations}

Lastly, we analyze the limitation of our approach. Note that it is not fair to directly compare the results by our language-only model and the deep visual feature-based models, as our approach uses additional annotations that deep visual features does not, i.e., annotated sentences from Localized Narratives and COCO captions. This means that the sentences describe the content prominent for human eyes. The questions in the dataset, also annotated by humans, are based on the same perspective, and our model may take this advantage. Meanwhile, deep visual features are trained in the end-to-end manner, which can provide strong supervision on what to see. This on the other hand benefits deep visual feature-based models. Yet, our results give interesting insights on differences and analogies between deep visual features and textual representation, providing a baseline for the VQA tasks with the interpretable representation. Moreover, this is study brings us the opportunity to introduce a new research direction for VQA in particular, and image understanding in general: to automatically generate image description as image representations instead of (or combined with) deep visual features.

\section{Conclusion}
In this paper, we explored using textual representations of images instead of deep visual features for VQA.
Additionally, we explored data augmentation methods for descriptions and questions to increase the training data size and diversity.  
Through the experiments, including ablations, we validated our language-only model competed with deep visual feature-based models. 
Also, most of the data augmentation techniques enhanced the performance; especially the boost by back translation for questions was outstanding. 
It was revealed that machines do not need to take a thousand words to understand an image, but only a hundred.
In this study, we dealt with VQA to validate the effectiveness of the textual representations, but these representations may be potentially applied to other tasks, not only VQA.
In future work, we will 1) delve into why back translation for questions works so well; 2) delve into the textual representations for various tasks; 3) explore the performance when using a text generator to make the image descriptions.

%%\bibliographystyle {plainnat}
\bibliographystyle{ACM-Reference-Format}
\bibliography{sample-base}

\end{document}